\documentclass[runningheads]{llncs}
\usepackage{caption}
\usepackage{graphicx}
\usepackage{array,xcolor,colortbl}
\usepackage{multirow}
\usepackage[outdir=./]{epstopdf}
\usepackage[export]{adjustbox}
\usepackage[margin=1in]{geometry}
\usepackage{siunitx}
\usepackage{array}
\usepackage{subfig}
\usepackage{color, colortbl}
\usepackage[section]{placeins}
\usepackage{newunicodechar}
\usepackage[utf8]{inputenc}
\usepackage{textcomp}
\usepackage{xcolor}
\usepackage{algorithmic}
\usepackage{pdfpages}
\usepackage{float}
\usepackage{placeins}
\usepackage{comment}
\usepackage{hyperref}
\hypersetup{colorlinks,allcolors=black}

\begin{document}
\title{Predicting Financial Literacy via Semi-supervised Learning}
\author{David Hason Rudd\inst{1} \and
Huan Huo\inst{1} \and
Guandong Xu\inst{1,2}}
\authorrunning{D. Hason Rudd et al.}
%
\institute{The University of Technology Sydney, 15 Broadway, Ultimo, Australia \and
Advanced Analytics institute (AAi), 15 Broadway, Ultimo, Australia\\
\email{\{david.hasonrudd@student, huan.huo, guandong.xu\}@uts.edu.au}\\}

\maketitle              

\begin{abstract}
Financial literacy (FL) represents a person's ability to turn assets into income, and understanding digital currencies has been added to the modern definition. FL can be predicted by exploiting unlabelled recorded data in financial networks via semi-supervised learning (SSL). Measuring and predicting FL has not been widely studied, resulting in limited understanding of customer financial engagement consequences. Previous studies have shown that low FL increases the risk of social harm. Therefore, it is important to accurately estimate FL to allocate specific intervention programs to less financially literate groups. This will not only increase company profitability, but will also reduce government spending. Some studies considered predicting FL in classification tasks, whereas others developed FL definitions and impacts. The current paper investigated mechanisms to learn customer FL level from their financial data using sampling by synthetic minority over-sampling techniques for regression with Gaussian noise (SMOGN). We propose the SMOGN-COREG model for semi-supervised regression, applying SMOGN to deal with unbalanced datasets and a nonparametric multi-learner co-regression (COREG) algorithm for labeling. We compared the SMOGN-COREG model with six well-known regressors on five datasets to evaluate the proposed models effectiveness on unbalanced and unlabelled financial data. Experimental results confirmed that the proposed method outperformed the comparator models for unbalanced and unlabelled financial data. Therefore, SMOGN-COREG is a step towards using unlabelled data to estimate FL level.
\end{abstract}

\section{Introduction}

Financial literacy (FL) is an essential skill in the modern world, and is mandatory for consumers operating in an increasingly complex economic society~\cite{worthington2006predicting}. Current economic conditions have raised significant concerns regarding Australian's financial security~\cite{worthington2006predicting}, particularly for those who lack the resources and skills to withstand downswings in the economy and take advantage of upswings. Several studies have determined there is a need not only for better understanding, but also to improve FL level. Individuals are generally responsible for various financial decisions, most importantly regarding retirement preparation and house financing. Previously studies~\cite{worthington2006predicting} confirmed the relationship between the complexity of these choices and increased stakes, and also highlights consequences from making financial decisions without sufficient FL. Therefore, effective financial management is a critical factor for any organization to achieve efficiency and success in the market~\cite{worthington2006predicting}. Lusardi \cite{lusardi2010financial} surveyed FL definition and effects, evaluating FL levels by asking volunteers four questions about compound interest, inflation, time value of money, and risk diversification. He showed that risk diversification was the most challenging question, with only 9\% of Australians giving the correct answer. 

The current paper assessed the proposed model's effectiveness on unbalanced financial network data to predict customer FL in a superannuation company. Measuring FL levels for millions of customers through a particular online survey in each financial period would be extremely time-consuming and expensive, hence SSL, which exploits a small portion of labelled data and a large amount of unlabelled data, is a smart approach. The dataset was built from customers' financial interactions data and labelled according to an online questionnaire similar to Lusardi's study~\cite{lusardi2010financial}. To our best knowledge, no previous study exploits a large amount of unlabelled data to predict FL level. We used a baseline algorithm in the self-training method with ensemble cross-validation to justify the baseline model's robustness on unbalanced dataset(s). We applied SMOGN to oversample values to enable predicting rare or uncommon data in the skewed dataset. Empirical results confirmed that the proposed SMOGN-COREG model outperformed all current models. Thus, including unlabelled examples via SSR methods improves prediction accuracy more than using only labelled data in supervised methods. 

The remainder of this paper is organised as follows: Section 2 reviews recent related FL studies and Section 3 discusses sampling methods and semi-supervised learning. Section 4 discusses specific methods employed in this analysis and Section 5 analyses gathered data and addresses each research question in turn. Finally, section 6 summarises and concludes the paper, and discusses implications for the findings in real-world applications.

\section{Related Work}

\subsection{ Financial literacy studies}

The financial literacy literature can be categorised in two ways. The first category explains different FL survey generation, and the second concentrates on measuring FL. Most of the research makes use of surveys to evaluate and predict the FL level. Several large-scale surveys have been conducted aimed at establishing the distribution and levels of FL.

Worthington~\cite{worthington2006predicting} examined FL across 924 individuals, largely students at 14 different colleges, and associated their scores with socioeconomic and demographic characteristics. They built a database from respondent answers to an 80 question survey  covering three main areas: mathematics literacy, money management skills, and financial competence. A logit model was employed to predict FL level effectively, but their model was most precise or accurate at predicting highest and lowest FL levels with enigmatic effects on intervening cells in the model. Their model depended on density functions, producing high-accuracy results only when predicting the lowest and highest FL levels, whereas the middle 60\% responses remained unpredictable~\cite{worthington2006predicting}. Experimental results showed that students with lower FL levels were likely to live in deprived areas, were unlikely to be business majors, and did not have much work experience. Holding all other factors equivalent or constant, older, higher educated, farm owners, business owners, and university educated respondents exhibited better FL.

Observed FL levels have decreased since the early surveys, but seem to vary between demographic and socioeconomic groups~\cite{lusardi2010financial}. Factors that influence FL level include gender, age, ethnicity, occupation, education, income, savings, and debt. FL prediction gave high scores for professional and highly educated people aged between 50 and 60; with lowest scores for unemployed females, and those who spoke English as a second language~\cite{lusardi2010financial}.

Huang \cite{huang2007financial} proposed a back propagation neural network (BPNN) to evaluate FL level. BPNNs comprising three distinct two-hidden layer networks were employed to model credit cards, loans, and superannuation on different datasets with approximately 900 examples. Their results confirmed BPNN capability to simulate FL with 92\% overall performance. 

Most previous FL studies collected labelled data from questionnaire surveys, whereas the current study predicted FL level on many recorded unlabelled data from customer financial activities with only a small number of labelled data acquisitions from an online questionary.

\subsection{Recent semi-supervised learning approaches } 

Semi-supervised learning uses unlabelled and labelled data in the learning process, in contrast with supervised and unsupervised learning methods, which use only labelled and unlabelled data, respectively. Having human annotators label data is prohibitively expensive and time-consuming, whereas unlabelled data acquisition for learning is easy and fast. However, although using unlabelled data via semi-supervised learning (SSL) is a good method to reduce human effort and improve model performance, some challenges make model tuning more time-consuming and critical than with other machine learning techniques.
Depending on the target variable type in model output, SSL can be categorised into two main approaches: semi-supervised regression (SSR) and semi-supervised classification (SSC). SSC is used where the target variable is discrete, whereas SSR is the better choice when model output is continuous.

Different SSL methods are used to fit the structure of a problem, such as maximising expectation with generative mixture models, self-training, co-training, transductive support vector machines (SVMs), and graph based methods~\cite{fazakis2019multi}. Ding et al.~\cite{ding2018semi} proposed GraphSGAN, applying SSL on graphs with generative adversarial networks. They experimentally confirmed the proposed approach on various datasets, including labelled and unlabelled datasets, performed significantly better than other methods, such as Chebyshev~\cite{defferrard2016convolutional} and graph convolutional networks (GCNs)~\cite{kipf2016semi}, and was more sensitive to labelled data~\cite{ding2018semi}.  GraphSGAN generated fake nodes in the density gap, reducing node influences across the density gap. Higher curvature for learned classification function around density gaps was achieved by discriminating fake from real samples. 
Lin and Gao ~\cite{9157520} proposed graph-based semi-supervised learning. They set up a shoestring framework using a typical graph based SSL, with two-layer graph convolutional neural network as a prototypical model for learning nonlinear mapping of nodes into an embedding vector, and then applied a metric learning network on the embedding vector to identify and learn pair-wise similarity between node and centroid representation in each class. The proposed method was tested on seven models and five datasets with 20 labelled data points in each class, achieving better classification performance than baseline methods.

Although the above approaches could theoretically adopt any current SSL methods, most were applied as supervised methods for classification, since real-valued target variables raise practical difficulties for SSL in regression. Motivated by these earlier studies, our methodological approach (see Section 4) is a mixed methodology SMOGN-COREG semi-supervised learning contribution to measuring FL.

\section{Preliminary Knowledge}

\subsection{Sampling techniques by SMOGN}

The unbalanced learning problem is concerned with learning algorithm performance in the presence of underrepresented data and severely skewed class distributions~\cite{vluymans2019learning}. The well-known synthetic minority oversampling technique for regression (SMOTER) extends the SMOTE algorithm for regression tasks and is commonly used in pre-processing to handle unbalanced datasets by generating synthetic samples for minority classes.  Torgo and Ribeiro~\cite{torgo2007utility} defined a relevance function to determine normal and rare value sets and map them onto a relevance scale between 0 and 1, representing minimum and maximum relevance, respectively. 

A threshold $t_R$ was established on relevance values assigned to each user to define the rare value set as 
$$
D_{R}=\left\{|x, y| \in D: \phi(y) \geq t_{R}\right\}
$$ 
and normal cases as 
$$ 
D_{N}=\left\{\langle x, y| \in D: \phi(y)<t_{R}\right\},
$$ 
where $D$ is a training set 
$$
\mathcal{D}=\left\{\left\langle\mathbf{x}_{i}, y_{i}\right\rangle\right\}_{i=1}^{N}
$$ 
with $N$ data points. The relevance function and $t_R$ are used to determine $D_R$ and $D_N$ sets in all sampling strategies. 

Branco et al. proposed SMOGN~\cite{branco2017smogn}, combining one random under-sampling and two oversampling techniques to increase data generation diversity, which cannot be achieved using only introduced Gaussian noise. SMOGN generates new synthetic data with SMOTER, which selects k-nearest neighbors (k-NN) based on the distance between two data points or introduces Gaussian noise~\cite{branco2017smogn}. SMOTER uses $t_R$ to determine whether neighbors are within safe or unsafe zones by calculating half the median distance between two data points. The main strategy is to classify important and less important cases in BinsR and BinsN partitions, and then apply oversampling and random under-sampling. 

\begin{figure}[!htb]
\centering
\includegraphics[scale=0.36]{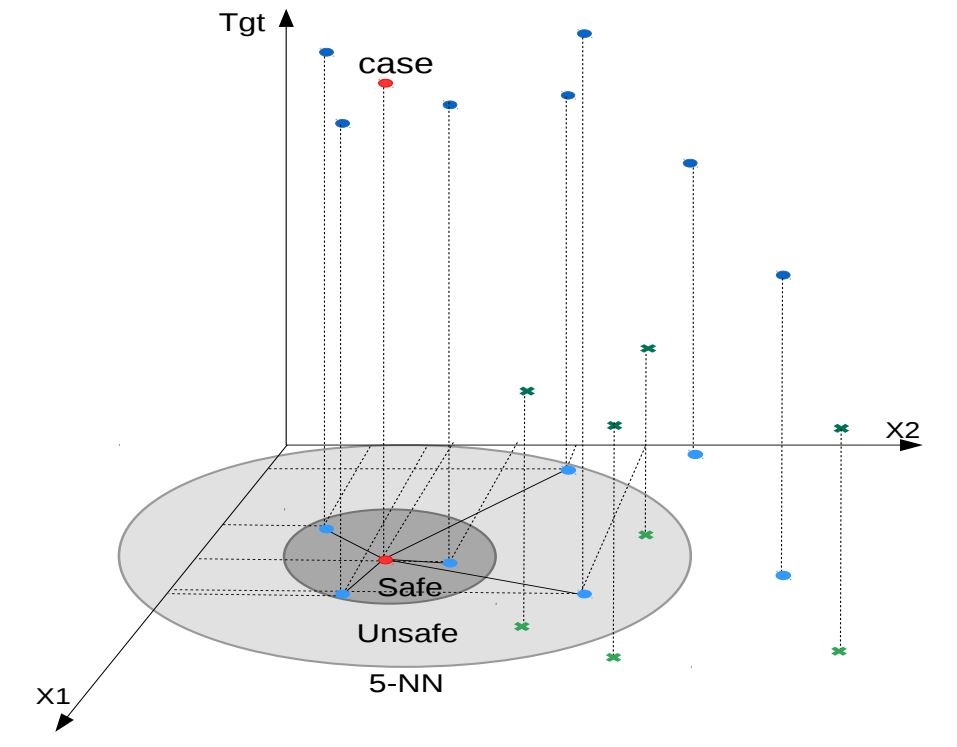}
\caption{Synthetic examples in SMOGN~\cite{branco2017smogn}}
\label{fig:f1}
\end{figure}

Figure~\ref{fig:f1} shows a SMOGN synthetic instance for seed cases with five nearest neighbors. Three neighbors are within the safe distance and the other two are at unsafe distance. This synthetic example shows that instances belonging to the normal bin (green) are more likely to overlap with instances associated with the relevant bin within the unsafe distance. Thus, SMOGN generates new synthetic examples and SMOTER selects K-NN or Gaussian noise based on the distance between the data points. If the neighbour is within a safe distance, it is suitable to conduct interpolation via the SMOTER method. On the other hand, if the selected neighbour is located in an unsafe zone introducing Gaussian Noise is a better selection to generate a new instance. 

\subsection{Co-regression semi-supervised learning}

Co-regression (COREG) \cite{zhou2005semi} is a nonparametric multi-view and multiple learner SSR method. The flexible COREG SSL algorithm implements two regressors, one labels unlabelled data for the other, and labeling confidence for unlabelled data is determined by the sum of the mean squared error reduction over the labelled neighborhood for that data point. Final prediction is made by averaging regression estimates generated by both regressors. COREG uses a lazy learning method including two k-NNs, which improves computational load because the lazy learner does not hold a separate training phase and refine regressors in each iteration. In contrast neural networks or regression trees require many labeling iterations, with consequently heavy computational load~\cite{zhou2005semi}.

COREG employs two k-NNs to compute mean squared error (MSE) for each $X_u$ to identify the most confidently labelled data by maximising 
\begin{equation} 
    \Delta_{\mathbf{x}_{u}}=\sum_{\mathbf{x}_{i} \in \Omega}\left(\left(\mathbf{y}_{i}-h\left(\mathbf{x}_{i}\right)\right)^{2}-\left(\mathbf{y}_{i}-h^{\prime}\left(\mathbf{x}_{i}\right)\right)^{2}\right) ,
    \label{equation_d}
\end{equation}
where  $h$ and $h^{'}$ are the original and refined k-NN regressors, respectively; and $\Omega$  is the set of k-NN labelled points from $X_U$. Information provided for regressors $\left(\mathbf{x}_{u}, \hat{\mathbf{y}}_{u}\right)$ by where $\hat{\mathbf{y}}_{u} $defines with  $\hat{\mathbf{y}}_{u}=h\left(\mathbf{x}_{u}\right)$.

\section{Methodology}

This section discusses proposed SMOGN-COREG semi-supervised learning model techniques for regression. When selecting an SSR algorithm for labeling unlabelled data, we need to consider an algorithm with low computational load and superior results on large unbalanced datasets.

\subsection{Problem Statement}

Depending on the model application, SSL can be categorised into inductive and transductive frameworks. Inductive semi-supervised learning can handle unseen data, whereas transductive learning only works on labelled 
$$
\left\{\left(\mathbf{x}_{i}, y_{i}\right)\right\}_{i=1}^{l} \stackrel{i i d}{\sim} p(\mathbf{x}, y)
$$ 
and unlabelled 
$$
\left\{\mathrm{x}_{i}\right\}_{i=l+1}^{l+u} \stackrel{i d}{\sim} p(\mathbf{x})
$$
training data, where $L$ and $U$ are labelled and unlabelled data, respectively; $X$ is an input data point, $y$ is a target label, $P(X, y)$ is the unknown joint distribution, $p(X)$ is marginal (typically $p(X) = l \ll u$). The transductive method is only interested in labelled data~\cite{kostopoulos2018semi} 
$$
\left\{\mathbf{x}_{i}\right\}_{i=l+1}^{l+u}.
$$

The proposed SSL method can be expressed as 
$$
X = (x_{i})_{i\in[n]},
$$
where $n$ is total number of instances and $x$ is independent predictor into labelled set $X_{l} =(x_{1},...,x_{l})$ associated with labelled data $Y_{l} = (y_{1},...,y_{l})$ and unlabelled instances $X_{u} =(x_{l}+1,...,x_{l}+_{u})$, where labelled data are not available. 

This paper focused on SSR because target value FL is continuous, and we employ SMOGN sampling on all datasets to balance target variable distributions.

\subsection{SMOGN-COREG semi-supervised regression }

We applied SMOGN for pre-processing before the learning phase to improve model performance. The SMOGN regression technique combines under-sampling the majority class (values usually found near the mean for a normal distribution in response variable y) and oversampling the minority class (rare values in a normal distribution of y, typically found at the tails). SMOGN uses a synthetic minority over-sampling technique for regression, with the additional step of using Gaussian noise to perturb interpolated values. SMOGN applies a function $\phi$ to the dependent variable, generating corresponding $\phi \in [0, 1]$ for each value to decide whether an observation is in the majority or minority depending on $t_R$ (defined in the arguments). 

\begin{figure}[!h]
\centering
\includegraphics[scale=0.057]{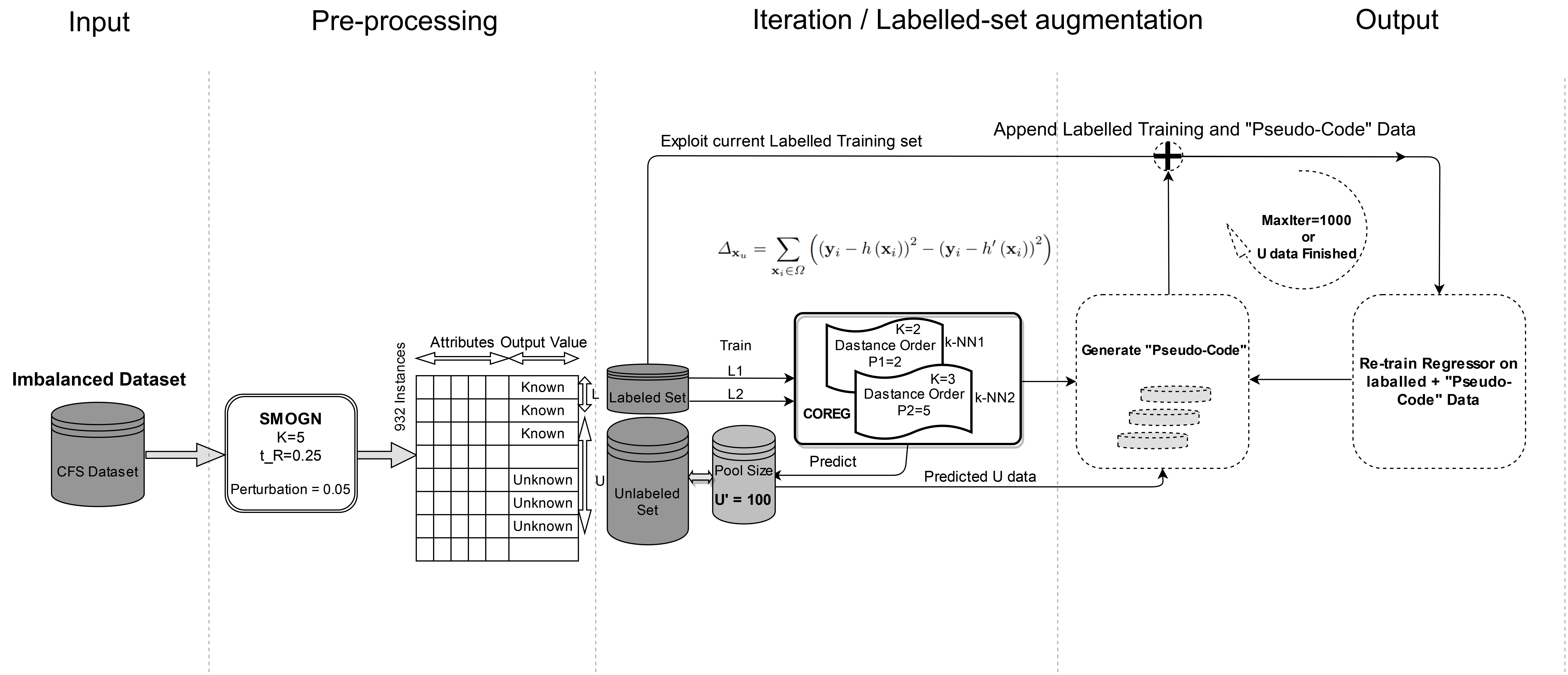}
\caption{Proposed SMOGN-COREG model workflow}
\label{fig:f2}
\end{figure}

Synthetic values in categorical features are created by randomly selecting observed values contained within their respective function. After post-processing, SMOGN returns an updated data frame with under and oversampled (synthetic) observations. Figure~\ref{fig:f2} show the model process workflow comprises four main steps: 
\begin{enumerate} 
\item input data; 
\item pre-processing: applied data cleansing, feature selection, and sampling;
\item labelled set data augmentation: semi-supervised regression; and 
\item output.  
\end{enumerate} 
The sampling strategies improved learning performance by increasing the small number of important rare cases.

We used COREG for different distance metrics rather than requiring sufficient and redundant views. COREG has broad applicability and can successfully use unlabelled data to boost regression predictions. Thus, combining sampling strategies with a non-parametric multi learner semi-supervised regression algorithm considerably improves performance on unbalanced datasets. Both base regressors are co-trained on the primarily labelled set with size 
\begin{equation}
    R=\frac{|L|}{|D|} , 
\label{equation_r}
\end{equation}

where $D=L \cup U$ is the total dataset, $|L| \ll|U|$;  $L$ is the initial training set, and $U$ denote the initial unlabelled set. 

Most focus for unbalanced domains is related to classification problems, whereas research into learning algorithms are less often explored to deal with unbalanced regression. Many important real-life applications, including the economy, crisis management, fault diagnosis, and meteorology, require predicting underrepresented data and important continuous target variables. Important rare cases often accompanied by a plethora of common values cause abnormal behavior for unbalanced learning scenarios~\cite{krawczyk2016learning}.

Unbalanced datasets in regression cause more difficulty than in classification because the number of values can be practically infinite for continuous target variables. Prediction performs more poorly when important data points are poorly represented, and the target variable is distributed on unequal user preferences compared to distributions with more frequent data points~\cite{branco2017smogn}.

\section{Experimental Procedure and Results}

\subsection {Datasets}
This study obtained five datasets obtained from a superannuation company for real-world experiments to verify the proposed model's effectiveness. The various datasets including considerable skewed data due to the high cardinality ratio. We intended to use balanced datasets and hence obtain reliable evaluation results. Table~\ref{tab:Datasets} lists the five unbalanced datasets and their attributes. CFS\_2017-2018\_FL contains 68 features (54 integer and 14 real variables) and 931 instances,  and the other four datasets contain with 89 features (54 integer, 16 polynomial, and 19 real variables) with approximately 900 labelled and unlabelled data points, for members holding accounts with Australian superannuation company Colonial First State (CFS). Dataset features included customer financial activities, demographics, income, account balance, marital status, age, employment, and some private features used in customer relationship management teams. We added the target variable "Financial literacy" in final dataset separately after Extract, Transform, Load (ETL) data from various sources, where FL value was derived from CFS online survey questionnaire in 2017 and 2018. However, the number of customers that participated in this survey was only approximately 5\% of the whole population, i.e., approximately 5\% of the data was labelled, and the rest remained unlabelled in all five datasets. 

\begin{table}[h!]
\centering
\scriptsize
\caption{Datasets collected}
\begin{tabular}{lccc} \hline
Dataset & \# Attributes & \# Instances & Size \\ \hline
CFS\_201706 & 89 & 824 & 73336 \\
CFS\_201712 & 89 & 856 & 76184 \\
CFS\_201806 & 89 & 899 & 80011 \\
CFS\_201812 & 89 & 918 & 81702 \\
CFS\_2017-2018\_FL & 68 & 931 & 64239 \\ \hline
\end{tabular}  
\label{tab:Datasets}
\end{table}

\subsection{Baseline and regression method configuration}

The regression methods were implemented using the Weka \footnote{Weka is a collection of machine-learning algorithms for data mining tasks in the Java SE platform, operating Windows, OS X, Linux} platform, and we compared proposed SMOGN-COREG model performance with the following supervised and semi-supervised models.

\begin{itemize}
    \item Linear Regression (LR) is the most popular method when there is a linear relationship between two features. We used Akaike information criterion (AIC) for model selection, and LR can deal with weighted instances. 
    \item k-NN using Euclidean distance, where $K = k \in \{4, 7, 9\}$.
    \item Sequential minimization optimization (SMOreg) to implement SVM with regression using a polynomial kernel with batch size = 100.
    \item M5 Rules model tree in if-then form, with minimum instances per leaf = 4.
    \item M5 Model Trees, a well-known model tree algorithm in Weka tools, constructs multivariate linear regression trees, with minimum instances per leaf = 4.
    \item Random Forest (RF) with tree depth = unlimited, and iteration and batch size = 100.
    \item Meta multi-scheme SSR algorithm (MSSRA)~\cite{fazakis2019multi} as the baseline model. MSSRA used three k-NN base regressors (3,7,9 k-NN), followed by self-training to enhance the labelled set by exploiting the unlabelled set and one final random forest regressor deployed for retraining after iteration. Labels for unknown test instances were then exported. Since the algorithm uses different regressors outside the iterative process, it can be considered a diversity booster, confirmed in the experiment by its robust results.
\end{itemize}
Several more supervised regressors were utilised for comparison purposes, but we did not include them the model comparison due to their considerably unsatisfactory performance.

\subsection{Experiment setup}

We initially employed cross-validation with 10 folds of the datasets, one fold for the test set and the remainder for learning. Unlabelled ratio UR = 80\% was used to split the training set in each fold, i.e., only 20\% labelled data were involved in learning. COREG maximum iterations = 100, $U' \text{ pool size} = 100$, and always $\Delta_{{x}_{u}}$ greater than zero in each iteration, hence maximum labeling capacity = 50000 iterations. However, this capacity was somewhat optimistic considering the negative impact from noisy data in the $L$ subset. Given the labeling capacity of the algorithm, all unlabelled data in the five datasets would be evaluated for labeling well before the maximum number of iterations was reached. Experimentally trading off between iterations and model runtime, we found 500 iterations and $U^{'}$ pool size = 100 unlabelled data was optimal and covered all confident predictions to enhance the labelled set during learning on all five datasets. 

We set the distance order for the two k-NN regressors in COREG as $K = 2$ and 3, respectively; with $K=2$ for the SMOGN algorithm oversampling, and $t_R = 0.25$. Gaussian noise introduced in SMOGN = 5\%, hence perturbation = 0.05 and maximum iteration = 1000. The pool contained 100 unlabelled examples randomly selected from the unlabelled set in each iteration. The final prediction outcome is the average regression predictions for the two regressors. Average MSE was recorded for labeling most confidence instances. 

We compared the proposed approach with one SSR algorithm and five widely used supervised regressors on five different datasets (see Table~\ref{tab:Datasets}). We considered four well-known evaluation metrics to determine regression performance:
\begin{itemize}
\item R-squared ($R^2$) (\ref{equation_R-squared}), 
\item Pearson correlation coefficient ($PCC$) (\ref{equation_pcc}), 
\item root mean squared error ($RMSE$) (\ref{equation_rmse}), and 
\item mean absolute error ($MAE$) (\ref{equation_mae});
\end{itemize}
Which can be expressed as 
\begin{equation}
    R^{2}=1-\frac{R S S}{T S S} , 
\label{equation_R-squared}
\end{equation}

\begin{equation}
    PCC =\frac{\sum_{i=1}^{n}\left(y_{i}-\bar{y}\right)\left(y_{i}^{\prime}-\bar{y}^{\prime}\right)}{\sum_{i=1}^{n}\left(y_{i}-\bar{y}\right)^{2}\left(y_{i}^{\prime}-\bar{y}^{\prime}\right)^{2}} , 
\label{equation_pcc}
\end{equation}

\begin{equation}
    RMSE =\frac{1}{n} \sqrt{\sum_{i=1}^{n}\left(y_{i}-y_{i}^{\prime}\right)^{2}} , 
\label{equation_rmse}
\end{equation}

\begin{equation}
    MAE =\frac{1}{n} \sum_{i=1}^{n}\left|y_{i}-y_{i}^{\prime}\right| , 
\label{equation_mae}
\end{equation}
respectively, where the dataset has $n$ values $\{y_1,...,y_n\}$ with $y_i$ values are real in a multivariate linear model with 
$$
Y_{i}=\beta_{0}+\sum_{j=1}^{P} \beta_{j} X_{i, j}+\varepsilon_{i}, 
$$ 
$y^{'}$ corresponding to the predicted value for data point $x_{i}$, $\bar{y}$ is the mean of the observed data; $\bar{y}, \bar{y}^{\prime}$ are mean values for $y_{i}$ and $y_{i}^{\prime}$, respectively. Larger $PCC$ and $R^2$ and smaller $MAE$ and $RMSE$ represent improved prediction accuracy. 

\subsection{Empirical Results} 

\begin{figure}[h!]
\centering
\includegraphics[scale=0.05]{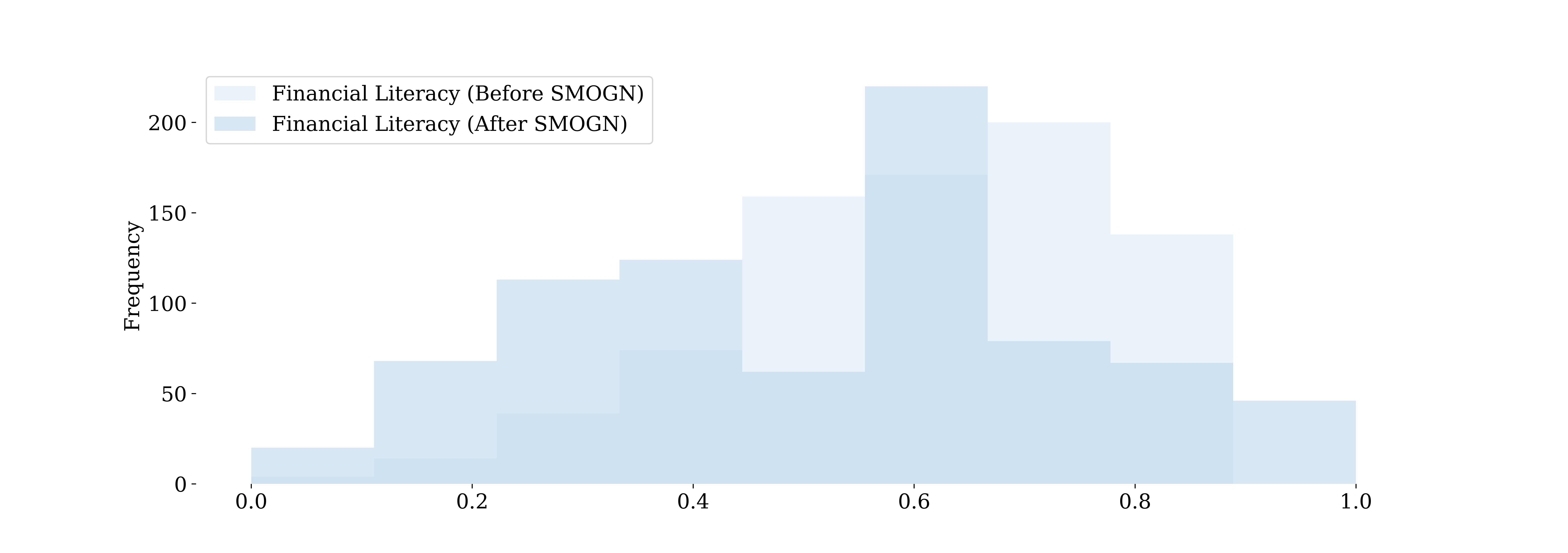}
\caption{Distribution of target variable before and after applying SMOGN on CFS\_2017-2018\_FL dataset}
\label{fig:f3}
\end{figure}

Based on the aim of achieving excellent results from SMOGN for tackling imbalanced regression problems, an experiment was conducted on the main dataset CFS\_2017-2018\_FL and its results presented in Figure \ref{fig:f3} show a clear pattern of the imbalanced target variable distribution, with a higher density over 0.5, and vice versa. The dark blue histogram shows that after applying SMOGN, fewer sample data points were extracted between the values of 0.6 and 0.9. In contrast, some extra samples were generated from lower than 0.5 values. Thus we can claim that the skewed data distribution was modified after applying SMOGN. On the other hand, this improvement in data distribution led to boosting the COREG learning capability to generate higher accuracy in minority samples. It concludes that the mixed methodology of oversampling and undersampling by the SMOGN and COREG is worked efficiently not only in our targeted dataset and other similar finance datasets.  

The experimental results on main dataset "CFS\_2017-2018\_FL" presented in the tables and graphs prove that the proposed model performed better than the other regression algorithms. Specific results of the aforementioned evaluation metrics on five different datasets are as follows:  

\begin{itemize}

    \item    
    The RMSE results shown in Table \ref{tab:RMSE}. are in an acceptable range for the proposed model with minimum 
    \begin{table}[!htb]
    \centering
    \scriptsize
    \caption{RMSE results of supervised regressors, baseline MSSRA and proposed SMOGN-COREG model}
    \begin{tabular}{p{0.25\textwidth}>{\centering}p{0.20\textwidth}>{\centering}p{0.12\textwidth}>{\centering}p{0.12\textwidth} >{\centering}p{0.12\textwidth}>{\centering\arraybackslash}p{0.12\textwidth}} \hline
    Datasets & CFS\_2017-2018\_FL & CFS\_201812 & CFS\_201806 & CFS\_201712 & CFS\_201706 \\\hline
    MSSRA & 0.1367 & 0.1565 & 0.1618 & 0.156 & 0.1549 \\
    SMOGEN-COREG & 0.1356 & 0.1335 & 0.1303 & 0.1285 & 0.1416 \\
    4-NN & 0.1344 & 0.1483 & 0.1581 & 0.1502 & 0.1513 \\
    7-NN & 0.1306 & 0.1439 & 0.153 & 0.1447 & 0.1465 \\
    9-NN & 0.1284 & 0.1426 & 0.1533 & 0.1448 & 0.1462 \\
    SMOreg & 0.1317 & 0.1321 & 0.1325 & 0.1831 & 0.1263 \\
    LR & \textbf{0.1275} & 0.1224 & 0.1304 & 1.1528 & 0.1251 \\
    M5 & 0.1276 & \textbf{0.1214} & \textbf{0.1223} & \textbf{0.1227} & \textbf{0.1207} \\
    M5rules & 0.1277 & 0.1215 & 0.1229 & 0.1231 & 0.1208 \\
    RF & 0.1317 & 0.1339 & 0.1359 & 0.1362 & 0.1361 \\\hline
    \end{tabular}
    
    \label{tab:RMSE}
    \end{table}
    
     RMSE 0.1285. The M5 had the best result with the lowest RMSE at 0.1207; in contrast, the LR indicated poor performance with the highest RMSE of 1.1528. On the other hand, the data analysis in Table \ref{tab:RMSE} showed that the lower fluctuation level is obtained in the proposed SMOGN-COREG and RF model with a standard deviation of 0.0051 and 0.0019, unlike the SMOreg and LR 0.023592117 and 0.459051896, respectively. This finding confirms additional evidence that although the RMSE result in the SMOGN-COREG is slightly higher than supervised models M5, SMOreg and LR  because the two SMOGN-COREG regressors train on the augmented training set built by the combination of the initial labelled set and Pseudo-code subset increased model prediction error in SSL in compare with supervised learning methods due to the inherent limitations of SSL, the amount of noise in data generate many incorrect pseudo-labels, leading to erroneous high confidence predictions. Despite this, the SMOGN-COREG algorithm's stability is higher than the baseline SSL algorithm, and other mentioned supervised learning models in the finance datasets.
  
    \item A significant improvement in R-squared and PCC values with our proposed model demonstrates the compatibility of the two algorithms, SMOGN and COREG, on the financial  dataset. Moreover, the results emphasise the importance of exploiting the sampling technique SMOGN to improve model performance on imbalanced datasets. The R-squared and PCC results in Figure 4. and Table \ref{tab:PCC}. show that the COREG algorithm achieved the lowest R-squared and PCC results at 0.4431 and 0.6656, respectively, in contrast to SMOGN-COREG, which obtains superior results of 0.7171 and 0.8468, respectively. 
    
    \begin{table}[!htb]
        \centering
        \scriptsize
        \caption{PCC results of supervised regressors, baseline MSSRA and proposed SMOGN-COREG model}
      \begin{tabular}{p{0.25\textwidth}>{\centering}p{0.20\textwidth}>{\centering}p{0.12\textwidth}>{\centering}p{0.12\textwidth}>{\centering}p{0.12\textwidth} >{\centering\arraybackslash}p{0.12\textwidth}} \hline
    Dataset & CFS\_2017-2018\_FL & CFS\_201812 & CFS\_201806 & CFS\_201712 & CFS\_201706 \\ \hline
    MSSRA (Base-model) & 0.7922 & 0.7465 & 0.7322 & 0.7476 & 0.7501 \\
    SMOGEN-COREG & \textbf{0.8468} & \textbf{0.8384} & \textbf{0.8622} & \textbf{0.8523} & 0.7454 \\
    4-NN & 0.7876 & 0.7274 & 0.6837 & 0.7171 & 0.7095 \\
    7-NN & 0.7988 & 0.7472 & 0.7117 & 0.7442 & 0.7325 \\
    9-NN & 0.806 & 0.754 & 0.7144 & 0.7461 & 0.7365 \\
    SMOreg & 0.7955 & 0.7919 & 0.7917 & 0.6295 & 0.8085 \\
    LR & 0.8092 & 0.8249 & 0.8025 & 0.0858 & 0.8134 \\
    M5 & 0.8094 & 0.827 & 0.8259 & 0.8225 & \textbf{0.8268} \\
    M5rules & 0.8092 & 0.8267 & 0.824 & 0.8214 & 0.8265 \\
    RF & 0.7985 & 0.8119 & 0.8081 & 0.809 & 0.8053 \\
    Improved \% & 4.6 & 1.4 & 4.4 & 3.6 & -8.8 \\ \hline
    \end{tabular}
    \label{tab:PCC}
    \end{table}

    \begin{figure}[h!]
    \centering
    \includegraphics[scale=0.25]{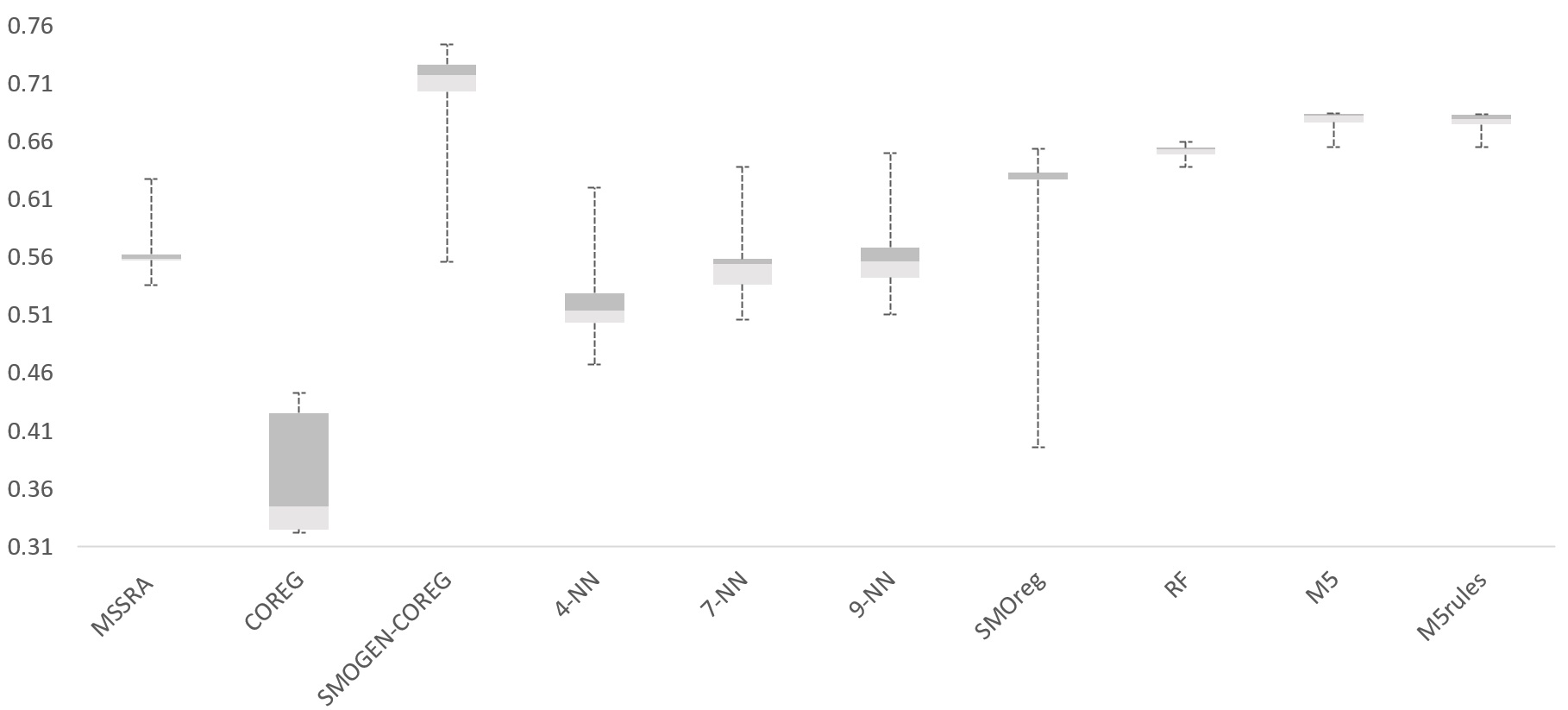}
    \caption{The proposed SMOGN-COREG model is obtained the best R-squared.}
    \label{fig:f4}
    \end{figure}

    \item The MAE values represented in the boxplot graph in Figure 5. show that M5 and M5Rules had a lower MAE than the other supervised- and semi-supervised-learning algorithms. RF had the least fluctuation in MAE, ranging from 0.102 to 0.104; however, the lowest MAE was achieved by the proposed model SMOGN-COREG, ranging from 0.0099 to 0.1091 on all datasets.
    
    \begin{figure}[!h]
    \centering
    \includegraphics[scale=0.24]{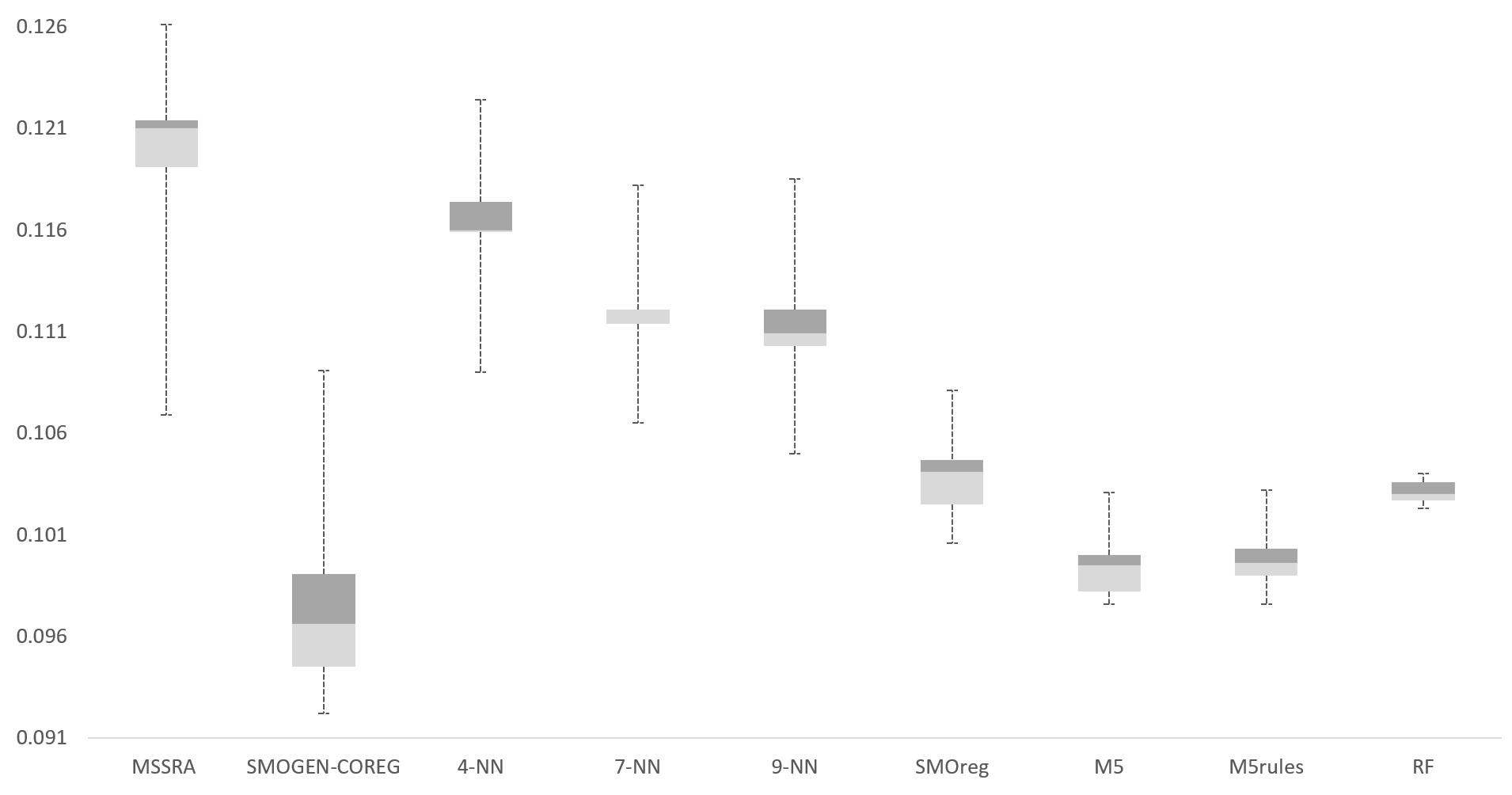}
    \caption{The SMOGN-COREG and M5 models achieved the better MAE result.}
    \label{fig:f5}
    \end{figure}

    \item In addition to the above results, it should be noted that these findings cannot be extrapolated to all type of datasets structures. The max cardinality of the dataset "CFS 201706" is significantly larger than other ones. Namely, the large number of unique values led to a low strength relationship between data points. As shown in RMSE results in Table 2. and PCC results in Table 3. the supervised model M5 performs better on the dataset "CFS 201706".
    
\end{itemize}

Therefore, from the final experiment, given the ability to predict the FL of customers, we can conclude two things: First, our hypothesis using a combination of sampling strategy techniques with a nonparametric multi-learner SSR algorithm provides better results than other regressors on imbalanced financial network data. Second, overall, exploiting a large amount of unlabelled data via SSR methods improves prediction accuracy more than using only labelled data in supervised methods on data collected based on customer financial activities. In Figure 5, the MAE in SMOGN-COREG improved 50\% over the COREG algorithm and 7.3\% over the baseline model. In Figure 4. SMOGN-COREG improved the R-squared 67\% and 18\% over the COREG and baseline algorithms, respectively.

The different ratio of the size of the unlabelled dataset to the total amount of data, called Unlabelled Ratio (UR), affects the R-squared, RMSE, and MAE. Figure 6. shows that the R-squared was imroved 31\% by increasing the number of unlabelled data points. The MAE was worse when learning was trained with a 70\% unlabelled ratio. Moreover, the RMSE value improved 9.8\% with a 99\% UR. Monitoring the model performance via the UR can be helpful for designers who want to simulate their own model output based on a different UR. The python implementation for the proposed model and result visualisation is available in GitHub (\url{https://github.com/DavidHason/predicting-financial-literacy}), to simplify reproducing and improving this study experiment results.

\begin{figure}[t!]
\centering
\includegraphics[scale=0.20]{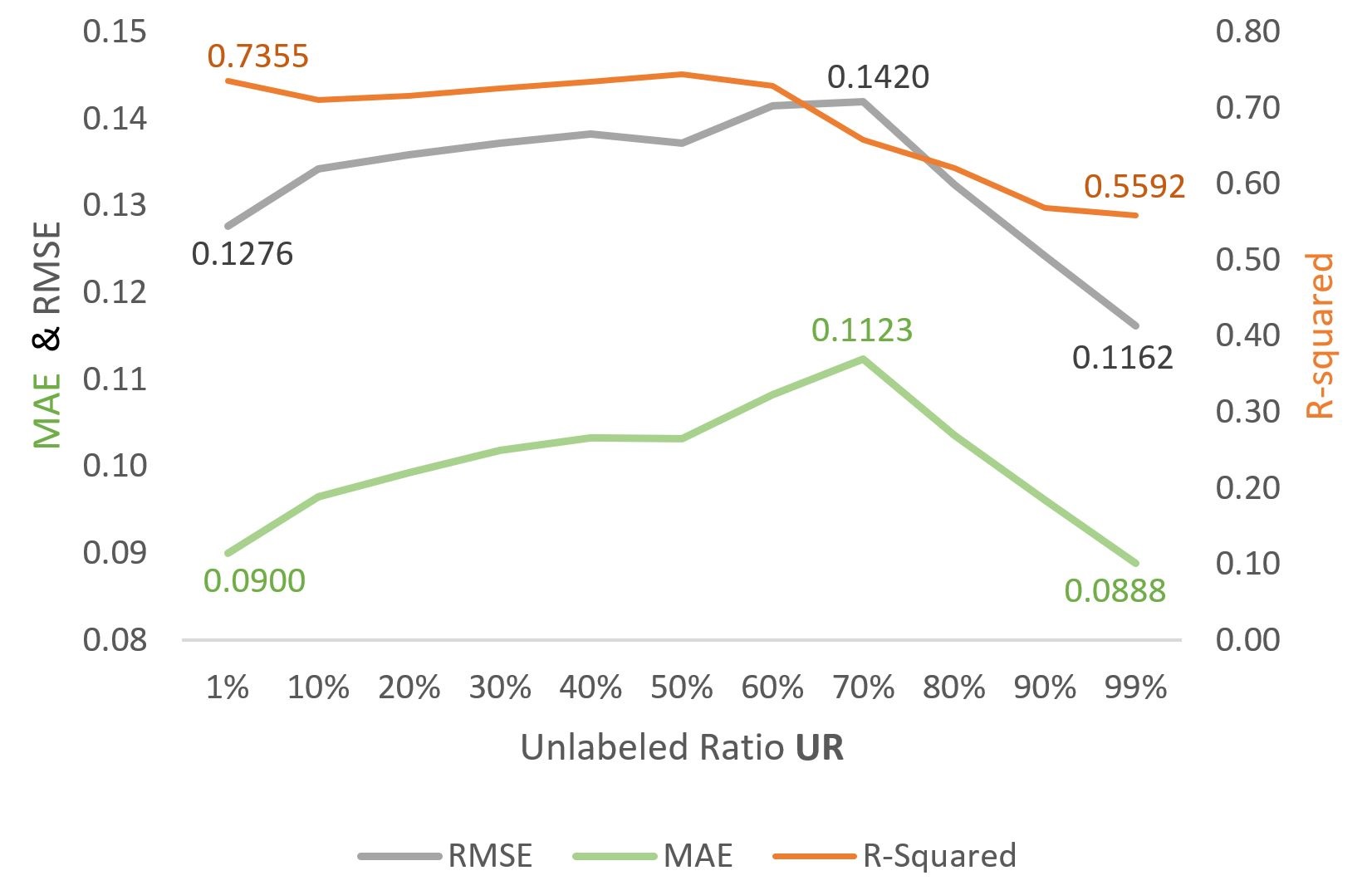}
\caption{The above graph shows the correlation of evaluation metrics results with UR.}
\label{fig:f6}
\end{figure}

\section{Conclusions}
Irrational financial decisions can have irreversible impacts on quality of life. Many relevant reports and articles have shown that people with poor FL are considerably more vulnerable to social harm and financial losses, such as job loss; family loss; reduced life expectancy; mental health problems; and most importantly, low-income retirement. To prevent this, it is essential to estimate FL and hence allocate specific intervention programs and financial advice to less financially literate groups. This will not only increase company profitability but also reduce government spending. Considerable financial data is recorded in the contemporary world, with a high proportion of that data unlabelled. Therefore, it is impossible to include this massive data repository in predictive models. The primary purpose for the current study was to develop a suitable method to predict FL level using financial datasets, which often include considerable unlabelled data.

Many SSL techniques have been used for various real-world applications, including GCNs, self-training, and co-training. Empirical results confirmed that combining SMOGN and COREG algorithms on unlabelled data reduced cost and model runtime, and improved prediction accuracy beyond current supervised regression methods. Most real-world problems involving unlabelled examples analysed using SSR methods have better prediction accuracy than using only labelled data in supervised methods. Thus, this study results represent a further step towards applying SSR techniques to assist FinTech companies in narrowing their consumer financial behavior and targeted marketing campaigns.

The proposed solution was based on an offline learning process because the proposed statistical predictive method was applied to previously collected data. Future study will investigate implementing online learning on streaming data. This would provide a predictive engine to conduct more accurate and trustworthy predictions, with additional data potentially coming from other financial institutions, such as other superannuation or FinTech companies. Using an active learning algorithm in the SSR models could be a potential method to achieve this. 

\section{Acknowledgement}
This work is partially supported by the Australian Research Council
(ARC) under Grant No. DP200101374 and LP170100891.

%

%
%
\bibliographystyle{splncs04}
\bibliography{mybibliography}
\end{document}